# Algebraic Statistics in Model Selection

Luis David Garcia
Mathematics Department
Virginia Polytechnic Institute and State University
Blacksburg, VA 24061
lgarcia@vbi.vt.edu

## Abstract

We develop the necessary theory in computational algebraic geometry to place Bayesian networks into the realm of algebraic statistics. We present an algebra–statistics dictionary focused on statistical modeling. In particular, we link the notion of *effective dimension* of a Bayesian network with the notion of *algebraic dimension* of a variety. We also obtain the independence and non–independence constraints on the distributions over the observable variables implied by a Bayesian network with hidden variables, via a generating set of an ideal of polynomials associated to the network. These results extend previous work on the subject. Finally, the relevance of these results for model selection is discussed.

## 1 INTRODUCTION

Two important aspects of model selection are the *effective dimension* of the model and the *observable constraints* that the model implies. Algebraically, the effective dimension of a model is the dimension of the associated variety and the observable constraints are the polynomials defining the variety. In this paper, we first present an algebra–statistics dictionary that allows us to translate concepts from one field to the other. This allows us to use powerful algebraic tools to better understand the concept of effective dimension. In particular, we provide several infinite families of models for which the effective dimension equals the *standard dimension* of the model, that is, the number of model parameters. Thus, no computations are required to obtain the effective dimension of a model belonging to these families. Finally, we use algebraic and geometric tools to obtain the independence and non–independence constraints on the distributions over the observable variables implied by a Bayesian network with hidden variables. For example, in (Garcia, Stillman and Sturmfels 2004) the authors applied this method to obtain the constraints implied by any *naive Bayesian network* with two hidden states, regardless of the number of levels of the observable variables. Moreover, these constraints are given in a compact syntactic representation. Previously, this result was restricted to the case of two binary features (see Geiger and Meek 1998).

A *Bayesian network*, also known as directed graphical model, is a family of probability distributions. In this paper, all random variables are assume to be discrete. In this case, these families are *real algebraic varieties* (actually, semi-algebraic sets): they are the zeros of some polynomials in the probability simplex (Geiger, Heckerman, King and Meek 2001; Settimi and Smith 2000). Bayesian networks can be described in two possible ways: parametrically, by an explicit mapping of a set of parameters to a set of distributions, or implicitly, by a set of independence constraints that the distributions must satisfy. This is an instance of the computer algebra principle that varieties can be presented either parametrically or implicitly (e.g., Cox, Little and O'Shea 1997).

The emerging field of *algebraic statistics* (Pistone, Riccomagno and Wynn 2001) advocates polynomial algebra as a tool in the statistical analysis of experiments and discrete data. In (Garcia, Stillman and Sturmfels 2004) the authors placed Bayesian networks into the realm of algebraic statistics, by developing the necessary theory in (computational) algebraic geometry and by demonstrating the effectiveness of polynomial algebra (Gröbner bases) for this class of models. This enabled the authors to express purely in algebro–geometric terms many concepts related to Bayesian networks. For example, the model known as *naive Bayesian network* corresponds to the higher secant variety of a Segre variety (see Catalisano, Geramita and Gimigliano 2002).

In this paper, we focus on the statistical concepts of *effective dimension* and *polynomial constraints* of Bayesian networks with hidden variables, with a view towards the problem of *model selection*. For instance, statisticians have known for decades that the dimension of a naive Bayes model can unexpectedly drop (e.g., Goodman 1974). The corresponding problem on secant varieties has been known for almost a hundred years (e.g., Terracini 1911). Recently, the importance of using the correct dimension of a model when applying the *Bayesian Information Criterion (BIC)* for Bayesian model selection was highlighted in (Rusakov and Geiger 2002; 2003). We use our algebro–geometric framework to shed light on some of the results obtained by these authors. In particular, we apply these algebraic tools to solve some of the empirical conjectures in (Rusakov and Geiger 2002; 2003 and Kočka and Zhang 2002) concerning effective dimension.

Moreover, we present new results concerning the independence and non–independence constraints on the distributions over the observable variables implied by a Bayesian network with hidden variables. These results can be used in model selection (e.g. Geiger and Meek 1998). In addition, these results are relevant on their own, since Bayesian networks with hidden variables are usually defined parametrically because the independence and non-independence constraints on the distributions over the observable variables are not easily established. This extends previous results obtained by (Geiger, et al. 2001; Geiger and Meek 1998).

## 2 MODEL SELECTION

Statisticians and other applied researchers are often faced with the problem of choosing the appropriate model that best fits a given set of observations. One possible approach to model selection is the Bayesian approach by which a model $M$ is chosen according to the maximum posteriori probability, given the observed data $D$:

$$P(M|D) \propto P(M,D) = P(M)P(D|M)$$

The probability $P(D|M)$ that the data $D$ is generated by the given model $M$ is obtained via integration over all possible parameter values with which the model is compatible. A closed–form formula for the *marginal likelihood* $P(D|M)$ is not known in general for Bayesian networks with hidden variables. But Schwarz (1978) derived an asymptotic formula for the marginal likelihood when the model is an affine subspace of the natural parameter space of an exponential family. This formula is known as the *Bayesian Information Criteria score*. Later on, Haughton (1988) established, under some regularity conditions, that the BIC score is a valid asymptotic rule for selecting models from a set of *curved exponential families*. In particular, this result holds for Bayesian networks without hidden variables, see (Geiger and Meek 1998). We note that although researchers have been using the BIC score for selecting models among Bayesian networks with hidden variables, it was shown in (Rusakov and Geiger 2002) that the BIC score is generally not valid for statistical models that belong to a *stratified exponential family*.

Bayesian networks with hidden variables are semi-algebraic sets, that is, the set of solutions of a finite number of polynomial equalities and inequalities. In terms of (Geiger and Meek 1998) these models are *stratified exponential families* (SEF), which is a class significantly richer than curved exponential families. There are two primary reasons for which the (standard) BIC score is in general not valid for this class. First, as mentioned before, the effective dimensionality of the model is no longer the number of network parameters (Geiger, et al. 2001). Moreover, the set of maximum likelihood points is sometimes a complex self–crossing surface. Algebraically, this means that the corresponding variety has singularities. An explicit example of a singular variety is given in Section 5.2. Finding effective solutions to these problems is crucial for Bayesian model selection. In (Rusakov and Geiger 2002; 2003) the authors developed a computational method to solve these problems by developing an *adjusted* BIC score.

## 3 ALGEBRAIC GEOMETRY OF BAYESIAN NETWORKS [1]

Let $X = \{X_1, \ldots, X_n\}$ be $n$ discrete variables having $r_1, \ldots, r_n$ states, respectively. We will further assume that each $X_i$ takes values in $[r_i] = \{1, 2, \ldots, r_i\}$. A Bayesian network model $M$ for variables $X$ is a set of joint distributions for $X$ defined by a graph $G_M$ and a set of local (multinomial) distributions $\mathcal{F}_M$. A probability distribution $P(x)$ belongs to the model $M$ if and only if it factors according to $G_M$ via

$$P(X = x) = \prod_{i=1}^{n} p_i(X_i = x_i \mid \text{pa}(X_i) = j), \quad (1)$$

where $x$ is an $n$-dimensional vector of values of $X$, $\text{pa}(X_i)$ denote the parents of node $X_i$ in $G_M$, $j$ denotes the values of $\text{pa}(X_i)$ in $x$ and $p_i$ is a conditional distribution from $\mathcal{F}_M$. We denote the *model parameters* defining $p_i(X_i = k \mid \text{pa}(X_i) = j)$ by $w_{ijk}$ and the

---
[1] This section follows very closely the introduction to Bayesian networks given in (Rusakov and Geiger 2003, §2) to accomplish a mirror effect and to unify notation.

*joint space parameters* $P(X = x)$ by $\theta_x$. The mapping that relates these parameters, derived from (1), is

$$\theta_{(x_1,\ldots,x_n)} = \prod_{i=1}^{n} w_{ijk}, \qquad (2)$$

where $k$ and $j$ denote the assignment to $X_i$ and $\mathrm{pa}(X_i)$ as dictated by $(x_1, \ldots, x_n)$. We can also think of the model parameters and the joint space parameters as algebraic *indeterminates*. This allows us to form two rings of polynomials $\mathbb{R}[\theta_x]$ and $\mathbb{R}[w_{ijk}]/J$. The ring $\mathbb{R}[\theta_x]$ will be our "ambient space". We will define *ideals* of this ring.

An ideal $I$ is a subset of a ring, which is closed under addition and multiplication by any polynomial in the ring. The interplay between algebra and geometry is given by considering the set of common zeros of all the polynomials in a given ideal $I$. This geometric object is called *algebraic variety* and is denoted by $V(I)$. For example, the circle $C \subset \mathbb{R}^2$ of radius 1 centered at the origin is given by the equation $x^2+y^2-1$. The intersection of $C$ with the line $y = x$ is given by the ideal $I$ generated by $\{x^2 + y^2 - 1, x - y\}$. For a nice introduction to the subject of *(computational) algebraic geometry* see (Cox, Little and O'Shea 1997).

The ring $\mathbb{R}[\theta_x]$ is generated by all indeterminates $\theta_x$ over the reals. The ring $\mathbb{R}[w_{ijk}]/J$ is generated by all indeterminates $w_{ijk}$, modulo the ideal $J$ generated by the relations

$$\sum_{k=1}^{r_i} w_{ijk} - 1.$$

Hence (2) induces a ring homomorphism

$$\Phi : \mathbb{R}[\theta_x] \to \mathbb{R}[w_{ijk}]/J.$$

Here, $J$ encodes the fact that $p_i(X_i = k \mid \mathrm{pa}(X_i) = j)$ is a probability distribution, for each fixed $i, j$. Algebraically, this means that any indeterminate of the form $w_{ijr_i}$ will be replaced by the linear form $1 - \sum_{k=1}^{r_i-1} w_{ijk}$. The ring $\mathbb{R}[\theta_x]$ is generated by $N = \prod_{i=i}^{n} r_i$ indeterminates. Note that $N - 1$ is the *complete dimension* of the model (Kočka and Zhang 2002). The ring $\mathbb{R}[w_{ijk}]/J$ is generated by

$$S = \sum_{i=1}^{n}(r_i - 1) \prod_{X_l \in \mathrm{pa}(X_i)} r_l$$

indeterminates. So, $S$ equals the *standard dimension* of the model (Kočka and Zhang 2002). The ideal consisting of all polynomials $f$ in $\mathbb{R}[\theta_x]$ such that $\Phi(f) = 0$ is denoted by $\ker(\Phi)$. It was proved in (Garcia, et al. 2004) that the intersection of the variety $V(\ker(\Phi)) \subset \mathbb{R}^N$ with the probability simplex

$$\Delta = \{(a_1, \ldots, a_N) \mid a_i \geq 0, \sum a_i = 1\}$$

is the set of all probability distributions in $M$. Note that computing generators for $\ker(\Phi)$ is the so-called *implicitization* problem (e.g., Geiger and Meek 1998).

The graph $G_M$ also describes the independencies of variables in $M$. The set of all independence relations encoded by $G_M$ is known as the set of *global Markov relations* (Lauritzen 1996). From this set we can construct an ideal $I_M$ in $\mathbb{R}[\theta_x]$. We will illustrate this with an example. Let $G_M$ be the following graph

$$X_3 \longrightarrow X_1 \longrightarrow X_2$$

The unique independence relation is $X_2 \perp\!\!\!\perp X_3 \mid X_1$, that is, $X_2$ is independent of $X_3$ given $X_1$. Observe that for this example we are working in the ring $\mathbb{R}[\theta_{(i_1,i_2,i_3)}]$ where $(i_1, i_2, i_3)$ represents an entry in a 3-dimensional $r_1 \times r_2 \times r_3$–table. For each $i \in [r_1]$ we will construct a $r_2 \times r_3$–matrix $A_i$ with entries in $\mathbb{R}[\theta_x]$. First index the rows of $A_i$ by $j \in [r_2]$ and the columns of $A_i$ by $k \in [r_3]$, then the $(j,k)$-entry of $A_i$ equals $\theta_{(i,j,k)}$, that is

$$A_i = \begin{pmatrix} \theta_{(i,1,1)} & \theta_{(i,1,2)} & \cdots & \theta_{(i,1,r_3)} \\ \theta_{(i,2,1)} & \theta_{(i,2,2)} & \cdots & \theta_{(i,2,r_3)} \\ \vdots & \vdots & \ddots & \vdots \\ \theta_{(i,r_2,1)} & \theta_{(i,r_2,2)} & \cdots & \theta_{(i,r_2,r_3)} \end{pmatrix}$$

The ideal $I_M$ is generated by all the $2 \times 2$–subdeterminants of the matrix $A_i$ for all $i \in [r_1]$. For instance,

$$\theta_{(1,1,1)}\theta_{(1,2,2)} - \theta_{(1,2,1)}\theta_{(1,1,2)}$$

is a generator of $I_M$. In this example, $\mathbb{R}[w_{ijk}]/J$ is generated by $(r_1-1)r_3$ indeterminates $w_{1jk}$, $(r_2-1)r_1$ indeterminates $w_{2jk}$, and $r_3 - 1$ indeterminates $w_{3k}$. The homomorphism $\Phi$ is given by

$$\theta_{(1,1,1)} \longrightarrow w_{111}w_{211}w_{31}$$
$$\theta_{(1,1,2)} \longrightarrow w_{121}w_{211}w_{32}$$
$$\vdots$$
$$\theta_{(r_1,r_2,r_3)} \longrightarrow \Big(1 - \sum_{i=1}^{r_1-1} w_{1r_3 i}\Big) \cdot \Big(1 - \sum_{j=1}^{r_2-1} w_{2r_1 j}\Big) \cdot$$
$$\Big(1 - \sum_{k=1}^{r_3-1} w_{3k}\Big)$$

Moreover, the ideal $\ker(\Phi)$ equals $I_M$.

Consider now the situation when some of the random variables in $M$ are hidden. After relabeling we may assume that the variables $X_{k+1}, \ldots, X_n$ are hidden, while the random variables $X_1, \ldots, X_k$ are observed. Thus, the *observable probabilities* are

$$\theta_{(i_1,\ldots,i_k,+,+,\ldots,+)} = \sum_{j_{k+1} \in [r_{k+1}]} \sum_{j_{k+2} \in [r_{k+2}]} \cdots \sum_{j_n \in [r_n]} \theta_{(i_1,i_2,\ldots,i_k,j_{k+1},\ldots,j_n)}.$$

We write $\mathbb{R}[\theta_{x'}]$ for the polynomial subring of $\mathbb{R}[\theta_x]$ generated by the observable probabilities. Let $\pi : \mathbb{R}^N \to \mathbb{R}^{N'}$ denote the canonical linear epimorphism induced by the inclusion of $\mathbb{R}[\theta_{x'}]$ into $\mathbb{R}[\theta_x]$. The following result appears in (Garcia, et al. 2004).

**Proposition 1.** *The set of all polynomial functions which vanish on the space of observable probability distributions is the prime ideal*

$$\ker(\Phi) \cap \mathbb{R}[\theta_{x'}].$$

We use this result in the next section to establish a connection between naive Bayesian models and secant varieties of Segre varieties. Moreover, in section 5, we use this result to compute the set of polynomial constraints on the distributions over the observable variables implied by a Bayesian network with hidden variables. This method is equivalent to the implicitization method proposed by Geiger and Meek (1998). The difference is that we compute the implicitization in two steps rather than one. First we compute the prime ideal $\ker(\Phi)$ corresponding to the model where all variables are assumed to be observed. Then we project the variety $V(\ker(\Phi))$ into the space of observable probability distributions

$$\pi\big(V(\ker(\Phi))\big) \subset \mathbb{R}^{N'}.$$

Usually, this method works faster than direct implicitization. More importantly, it enabled us to find a clear syntactic structure of the constraints implied by each Bayesian network studied in this paper.

## 4 NAIVE BAYESIAN NETWORKS AND EFFECTIVE DIMENSION

In this section we study the network $M$ which has $n+1$ random variables $X_1, \ldots, X_n, H$ and $n$ directed edges $H \to X_i$, $i = 1, 2, \ldots, n$. This is the *naive Bayesian network*. The variable $H$ is the hidden variable, and its levels $1, 2, \ldots, r$ are called the *classes*. The observed random variables $X_1, \ldots, X_n$ are the *features* of the model. For this model, the prime ideal $\ker(\Phi)$ coincides with the ideal $I_M$ (Garcia, et al. 2004) which

is specified by requiring that, for each fixed class, the features are completely independent:

$$X_1 \perp\!\!\!\perp X_2 \perp\!\!\!\perp \cdots \perp\!\!\!\perp X_n \mid H.$$

This ideal is obtained as the kernel of the map $\theta_{(i_1,i_2,\ldots,i_n,k)} \mapsto x_{i_1} y_{i_2} \cdots z_{i_n}$, one copy for each fixed class $k$, and then adding up these $r$ prime ideals. Equivalently, $\ker(\Phi)$ is the ideal of the *join* of $r$ copies of the *Segre variety*

$$V_{r_1,r_2,\ldots,r_n} = \mathbb{P}^{r_1-1} \times \mathbb{P}^{r_2-1} \times \cdots \times \mathbb{P}^{r_n-1} \subset \mathbb{P}^{r_1 r_2 \cdots r_n - 1}. \quad (3)$$

To familiarize ourselves with this notation, we work on the right–most inclusion for the case of two projective spaces. The embedding of $\mathbb{P}^{r_1-1} \times \mathbb{P}^{r_2-1}$ into $\mathbb{P}^{r_1 r_2 - 1}$ is defined (in homogeneous coordinates) by

$$\big((a_1,\ldots,a_{r_1}),(b_1,\ldots,b_{r_2})\big) \longmapsto \\ (a_1 b_1, a_1 b_2, \ldots, a_1 b_{r_2}, a_2 b_1, \ldots, a_{r_1} b_{r_2}).$$

Note that the variety $V = V_{r_1,r_2,\ldots,r_n}$ is contained in a space of dimension $r_1 r_2 \cdots r_n - 1$, which equals the complete dimension of the model $M$. Moreover, the dimension of $V_{r_1,r_2,\ldots,r_n}$ equal the dimension of $\mathbb{P}^{r_1-1} \times \mathbb{P}^{r_2-1} \times \cdots \times \mathbb{P}^{r_n-1}$ which equals the sum

$$d = r_1 + r_2 + \cdots + r_n - n.$$

We define a new variety in the following way. Take any two points in $V$ and consider the line spanned by them. Now take the closure of the union of all those lines. The result is a variety called the *secant variety of $V$*, denoted as $V^2$. In general, the $(r-1)$st *higher secant variety* of $V$ is defined as the closure of the union of all linear spaces spanned by $r$ points of $V$ and is denoted $V^r$. For an introduction to the subject see (Catalisano et al. 2002; Harris 1992).

The following result in (Garcia, et al. 2004) ties the algebro–geometric concept of secant varieties of Segre varieties with the statistical concept of naive Bayesian networks.

**Proposition 2.** *The naive Bayesian network $M$ with $r$ classes and $n$ features corresponds to the $r$-th secant variety of a Segre product of $n$ projective spaces:*

$$\overline{\pi\big(V(\ker(\Phi))\big)} = V^r = (V^r_{r_1,r_2,\ldots,r_n}) \quad (4)$$

This proposition states that the model $M$ equals the intersection of the variety $V^r$ with the probability simplex $\Delta$. There is an *expected dimension* for $V^r$. Recall that $\dim V = d = r_1 + r_2 + \cdots + r_n - n$, so one "expects" that

$$\dim V^r = \min\{N-1 = \prod_{i=1}^{n} r_i - 1, \quad rd+r-1\}, \quad (5)$$

where the number $rd+r-1$ corresponds to the choice of $r$ points on $V$ (which is $d$ dimensional), plus the choices of a point on the projective space $\mathbb{P}^{r-1}$ spanned by the $r$ points. When this number is too big, we should just get $\dim V^r = N-1$ (the dimension of the ambient space). We take a closer look at the number $rd+r-1$:

$$rd+r-1 = r(\sum_{i=1}^{n} r_i - n) + r - 1 = r\sum_{i=1}^{n}(r_i - 1) + r - 1$$

Hence $rd+r-1$ is the *standard dimension* of the model $M$. Thus, the expected dimension is defined as the minimum of the complete and the standard dimensions. Therefore, the *effective dimension* of $M$ is bounded by the expected dimension of $V^r$. Even more is true, the effective dimension of $M$ equals the dimension of $V^r$.

Let $M$ be the naive Bayesian network

$$X_1 \longleftarrow X_3 \longrightarrow X_2$$

with $r$ classes and 2 features. By (4), this model corresponds to the variety $V^r$, where

$$V = \mathbb{P}^{r_1-1} \times \mathbb{P}^{r_2-1} \subset \mathbb{P}^{r_1 r_2 - 1}.$$

It is a standard fact that the points on $V$ represent $r_1 \times r_2$–matrices of rank 1 (see Harris 1996). Moreover, $V^r$ consists of all real $r_1 \times r_2$–matrices of rank at most $r$. This variety is completely understood. In particular, the dimension of $V^r$ equals $r(r_1 + r_2) - r^2$. The dimension formula for this model obtained by Settimi and Smith (1998, Thm. 1) follows immediately. In the general case, note that for any $1 \le k \le n$, we have the following embeddings:

$$\mathbb{P}^{r_1-1} \times \cdots \times \mathbb{P}^{r_n-1} \longrightarrow (\mathbb{P}^{r_1} \times \cdots \times \mathbb{P}^{r_k}) \times (\mathbb{P}^{r_{k+1}} \times \cdots \times \mathbb{P}^{r_n})$$

$$\downarrow$$

$$\mathbb{P}^{\prod_{i=1}^{k} r_i - 1} \times \mathbb{P}^{\prod_{j=k+1}^{n} r_j - 1}$$

Since the previous maps are all inclusions, we get that the dimension of $V$ is bounded by the dimension of $U = \mathbb{P}^{\prod_{i=1}^{k} r_i - 1} \times \mathbb{P}^{\prod_{j=k+1}^{n} r_j - 1}$. Hence the dimension of $V^r$ is bounded by the dimension of $U^r$. Moreover, we can easily compute the dimension of $U^r$ using the formula obtained in the last paragraph. Of course, for each possible inclusion of $V$ into the product of two spaces $U$ we get a bound for the dimension of $V^r$. Denote by $dp$ the minimum among all these numbers. Theorems 3 and 4 in (Kočka and Zhang 2002) follows immediately, since we have shown that the effective dimension of $M$ is bounded by $dp$ and by the expected dimension.

Kočka and Zhang (2002) proposed the previous result as a bound for the effective dimension of $M$. Nevertheless, there is an infinite number of varieties $\mathbb{P}^{r_1-1} \times \cdots \times \mathbb{P}^{r_n-1}$ for which the number $dp$ is greater than the expected dimension. Hence, this upper-bound does not address the central question in this subject: When is the effective dimension of $M$ strictly less than the expected dimension of $M$? In algebraic geometry, we call such varieties *defective*. In fact, Lemma 3.1 in (Kočka and Zhang 2002) is a step towards answering this question. This is where our algebro–geometric formalism starts paying off. This question has a long history in algebraic geometry. For a nice introduction to the subject and many results see (Catalisano, et al. 2002). Here, we include some of their results.

**Catalisano, et al. 2002, Proposition 2.3.** *Let $M$ be the naive Bayesian network with $r$ classes and $n$ features denoted by $(r : r_1, r_2, \ldots, r_n)$. Then*

(i) *if $n = 2$ and $r = \min(r_1, r_2)$ then the effective dimension of $M$ equals the complete dimension $r_1 r_2 - 1$.*

(ii) *for $n = 2$ and $r < \min(r_1, r_2)$ the effective dimension of $M$ equals $r(r_1 + r_2) - r^2 - 1$.*

(iii) *for $n \ge 3$ and $r \le \min\{r_i \mid i = 1, \ldots, n\}$. the effective dimension of $M$ equals the standard dimension $r(r_1 + \cdots + r_n - n + 1) - 1$.*

**Catalisano, et al. 2002, Proposition 3.3.** *Let $M$ be the naive Bayesian network $(r : r_1, \ldots, r_n)$. Assume that $r_1 \le r_2 \le \cdots \le r_n$ and let $r$ be such that*

$$\prod_{i=1}^{n-1} r_i - \sum_{i=1}^{n-1}(r_i - 1) + 1 \le r \le \min\{r_n, \prod_{i=1}^{n-1} r_i - 1\}.$$

*Then $V^r$ is defective.*

**Catalisano, et al. 2002, Proposition 3.7.** *Let $M = (r : r_1, \ldots, r_n)$, with $n \ge 3$. We can assume that $r_n$ is the maximum among the $r_i$ (by relabeling the variables). If*

$$\left\lceil \frac{\sum_{i=1}^{n} r_i - n + 1}{2} \right\rceil \ge \max(r_n, r).$$

*Then the effective dimension of $M$ equals the standard dimension.*

In practice, the best method to compute the effective dimension of a model is the one implemented in (Rusakov and Geiger 2003). Nevertheless, the theoretical results obtained through our approach give infinite families for which such a computation is not needed, as shown by the previous propositions. Moreover, any attempt to obtain a tight bound on dimension or a classification of defective varieties would benefit from the algebraic geometry formalism provided by our approach.

As we mentioned earlier, the BIC score is generally not valid for Bayesian networks with hidden variables. The main reason is that the set of maximum likelihood points is sometimes a complex self–crossing surface, i.e., the corresponding variety has singularities.

Recently, Rusakov and Geiger (2003) provided an algorithm for analytic asymptotic evaluation of the marginal likelihood of data given a Bayesian network with hidden nodes. The algorithm requires the computation of the *deepest singularities* of the model at a single maximum likelihood model parameter. Then, it uses resolution of singularities to obtain the desired result. As noted by the authors, these computations become prohibitively large. They stand in contrast with the results obtained in their previous paper (Rusakov and Geiger 2002) where they were able to give a formula to compute the marginal likelihood for the special case of naive Bayesian networks consisting of binary variables and two hidden states. What allowed them to derive such a formula is that the singular locus of a binary naive Bayesian network had been previously computed in (Geiger et al. 2001). In particular, Geiger, et al. (2001) showed that the deepest singular locus $S'$ consists of all distributions where all variables are mutually independent and independent of the class node as well.

As we saw earlier, the naive Bayes model $M$ with two classes corresponds to the (first) secant variety of a Segre variety $V^2$. It is a standard fact in algebraic geometry that the $r$-th secant variety of any projective variety is always singular along the $(r-1)$-st secant variety. In our case, this means the naive Bayes model $M$ with two features is singular along the Segre variety $V$. Moreover, the variety $V$ corresponds to the model of complete independence $S'$ (see Garcia, et al. 2004). This extends the previously stated result in (Geiger, et al. 2001).

## 5 POLYNOMIAL CONSTRAINTS

Bayesian networks with hidden variables are usually defined parametrically because the independence and non–independence constraints on the distributions over the observable variables are not easily established. Since these constraints vary from one model to another they can be used to distinguish between models. Moreover, since these constraints are over the observable variables, their fit to data can be measured directly with some specially–designed statistical tests. For instance, the so–called *tetrad difference constraints* have been used for model selection and evaluation (see Spirtes, Glymour and Scheines 1993).

A step towards computing the constraints of a Bayesian network with hidden variables was given in (Garcia, et al. 2004), where the authors conjecture that any naive Bayes model $M$ with $r = 2$ classes is generated by the $3 \times 3$–subdeterminants of any two–dimensional table obtained by flattening the $n$-dimensional table $\theta_{(i_1, i_2, \cdots, i_n)}$. This conjecture was proved later on by Landsberg and Manivel (2003). The result concerning a naive Bayesian model with 2 classes and 2 ternary features obtained in (Geiger and Meek 1998) is one instance of this theorem.

We have computed the constraints on the distribution over the observable variables of all Bayesian network with three observable variables and one hidden variable. These results show that all but one of these models correspond to intersections (or joins) of (higher secant varieties of) Segre varieties. For some networks, we had to assume that the hidden variable is binary for the results to hold. The complete list of results will be given in a forthcoming paper. Here, we give a few examples including the one whose variety is not derived from a Segre variety. To simplify notation we will set $\theta_{i_1 \ldots i_n} = \theta_{(i_1, \ldots, i_n)}$.

### 5.1 A NETWORK WITH QUADRATIC CONSTRAINTS

Let $M$ be the following Bayesian network, where $X_4$ is hidden:

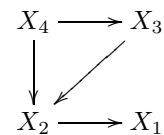

In this case we have $\ker(\Phi) = I_M = I_{1 \perp\!\!\!\perp \{3,4\}|2}$. Then $\ker(\Phi)$ is generated by the $2 \times 2$–subdeterminants of the $r_2$ matrices (see Garcia, et al. 2004)

$$A_j = \begin{pmatrix} \theta_{1j11} & \ldots & \theta_{1j1r_4} & \theta_{1j21} & \ldots & \theta_{1jr_3r_4} \\ \vdots & & \vdots & \vdots & & \vdots \\ \theta_{r_1j11} & \ldots & \theta_{r_1j1r_4} & \theta_{r_1j21} & \ldots & \theta_{r_1jr_3r_4} \end{pmatrix}.$$

Moreover, the ideal $Q_M = \ker(\Phi) \cap \mathbb{R}[\theta_{x'}]$ is generated by the $2 \times 2$–subdeterminants of the $r_2$ matrices

$$A_{j+} = \begin{pmatrix} \theta_{1j1+} & \theta_{1j2+} & \ldots & \theta_{1jr_3+} \\ \vdots & \vdots & & \vdots \\ \theta_{r_1j1+} & \theta_{r_1j2+} & \ldots & \theta_{r_1jr_3+} \end{pmatrix}.$$

Hence the model $M$ corresponds to the Segre variety $\mathbb{P}^{r_1-1} \times \mathbb{P}^{r_3-1}$.

*Proof.* First, note that each column of the matrix $A_{j+}$ is obtained by taking the sum of the corresponding $r_4$ columns of $A_j$. Denote by $I$ the ideal of all the $2 \times 2$–subdeterminants of the $r_2$ matrices $A_{j+}$. It is not hard to check directly that $I \subset Q_M$. Moreover, $Q_M$ is a prime ideal since $\ker(\Phi)$ is prime. On the other hand, $I$ is also a prime ideal since $V(I)$ is an irreducible variety. To see this observe that $V(I)$ is the join of $r_2$ disjoint irreducible varieties, each described by the $2\times 2$–subdeterminants of a generic matrix. The result follows from (Harris 1996, Thm. 11.14).

To show equality it suffices to show that $V(I) \subset V(Q_M)$. Let $p \in V(I)$, then $p$ is a $r_1 \times r_2 \times r_3$–matrix such that each slice $p_{j_0} = (p_{ij_0k})$ has rank 1. Let $P$ be the $r_1 \times r_2 \times r_3 \times r_4$–matrix defined by $P_{ijkl} = p_{ijk}/r_4$. For each slice $P_{j_0}$ consider the flattening $P'_{j_0} = (P_{ij_0kl})$, where the rows are indexed by $i \in [r_1]$ and the columns are indexed by pairs $(k,l) \in [r_3] \times [r_4]$. Clearly, $P'_{j_0}$ has rank 1, so $P \in V(\ker(\Phi))$. Thus $p = \pi(P) \in V(Q_{14})$. □

### 5.2 A NETWORK WITH CUBIC CONSTRAINTS

Let $M$ be the following Bayesian network where, $X_4$ is hidden:

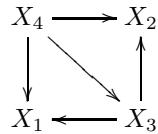

We have, $\ker(\Phi) = I_M = I_{1 \perp\!\!\!\perp 2 | \{3,4\}}$. Hence,

$$Q_M = \ker(\Phi) \cap \mathbb{R}[\theta_{x'}] = I_{1 \perp\!\!\!\perp 2|\{3,4\}} \cap \mathbb{R}[\theta_{x'}].$$

We will assume that the hidden variable is binary. Observe that

$$I = I_{1 \perp\!\!\!\perp 2|\{3,4\}} = \sum_{k_0 \in [r_3]} I_{k_0},$$

where $I_{k_0}$ is the ideal generated by the $2 \times 2$–subdeterminants of the following two $r_1 \times r_2$–matrices ($l \in \{1,2\}$ is fixed for each matrix).

$$\begin{pmatrix} \theta_{11k_0l} & \ldots & \theta_{1r_2k_0l} \\ \vdots & & \vdots \\ \theta_{r_11k_0l} & \ldots & \theta_{r_1r_2k_0l} \end{pmatrix}.$$

Moreover, the variety $V(I_{k_0} \cap \mathbb{R}[\theta_{x'}])$ equals the variety of all $r_1 \times r_2$–matrices of rank at most 2, given by the $3 \times 3$–subdeterminants of the $r_1 \times r_2$–matrix

$$\begin{pmatrix} \theta_{11k_0+} & \ldots & \theta_{1r_2k_0+} \\ \vdots & & \vdots \\ \theta_{r_11k_0+} & \ldots & \theta_{r_1r_2k_0+} \end{pmatrix}. \quad (6)$$

This is the secant variety of the Segre variety $\mathbb{P}^{r_1-1} \times \mathbb{P}^{r_2-1}$. Finally, we conclude that

$$Q_M = \sum_{k \in [r_3]} \left( I_k \cap \mathbb{R}[\theta_{x'}] \right).$$

This equality is obtained by showing that both ideals are prime, one is contained in the other and both have the same dimension. In particular, we obtain the effective dimension of this model:

$$\dim(M) = \dim(Q_M) = 2r_1r_3 + 2r_2r_3 - 4r_3.$$

Hence, this model is the join of $r_3$ secant varieties of Segre varieties $\mathbb{P}^{r_1-1} \times \mathbb{P}^{r_2-1}$. The polynomial constraints are given by the $3 \times 3$–subdeterminants of the $r_3$ matrices of the form (6), for each $k_0 \in [r_3]$.

### 5.3 A NETWORK WITH SEXTIC CONSTRAINTS

Finally, let $M$ be the following Bayesian network, where $X_4$ is hidden:

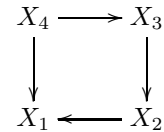

In this case we have

$$I_M = I_{1 \perp\!\!\!\perp 3|\{2,4\}} + I_{2 \perp\!\!\!\perp 4|3}.$$

This ideal is not radical in general so $I_M \subsetneq \ker(\Phi)$. If $r = 2$, i.e., $X_4$ is binary, then

$$Q_1 = I_M \cap \mathbb{R}[\theta_{x'}] = I_{1 \perp\!\!\!\perp 3|\{2,4\}} \cap \mathbb{R}[\theta_{x'}].$$

So $V(Q_M)$ is an irreducible proper subvariety of the irreducible variety $V(Q_1)$.

Consider the particular case $r_1 = r_4 = 2$ $r_2 = r_3 = 3$. First observe that the ideal $Q_1 = 0$. Moreover, the ideal $Q_M$ is generated by 3 irreducible *sextic* polynomials. The construction of these sextics goes as follows. Each pair $j_1, j_2 \in [r_2]$ specifies two matrices $N_{j_1}$ and $N_{j_2}$.

$$N_{j_1} = \begin{pmatrix} \theta_{1j_11+} & \theta_{1j_12+} & \theta_{1j_13+} \\ \theta_{2j_11+} & \theta_{2j_12+} & \theta_{2j_13+} \end{pmatrix},$$

$$N_{j_2} = \begin{pmatrix} \theta_{1j_21+} & \theta_{1j_22+} & \theta_{1j_23+} \\ \theta_{2j_21+} & \theta_{2j_22+} & \theta_{2j_23+} \end{pmatrix}.$$

The irreducible sextic polynomial arising from these matrices is given by

$$\theta_{+j_11+} U_1 V_1 - \theta_{+j_12+} U_2 V_2 + \theta_{+j_13+} U_3 V_3,$$

where the polynomial $U_s$ is the determinant of the $2 \times 2$–matrix obtained by eliminating the $s$-th column of $N_{j_1}$. The polynomial $V_s$ is the determinant of the $2 \times 2$–matrix $N'_{j_2}$ where the first column of $N'_{j_2}$ equals the $s$-th column of $N_{j_2}$ and the second column of $N'_{j_2}$ is the product of the remaining two columns of $N_{j_2}$. A similar result holds for any model of the form $(2, r_2, r_3, 2)$. We have conjectured that in fact one can extend this result to any model of the form $(r_1, r_2, r_3, 2)$, where one would obtain a sextic polynomial for each quadruple $i_1, i_2 \in [r_1]$ and $j_1, j_2 \in [r_2]$.

## Acknowledgments

This research was partially supported by NSF grant DMS–0138323. The author thanks Rebecca Garcia, Reinhard Laubenbacher, Christopher Meek and Bernd Sturmfels for helpful discussions. The author also thanks the anonymous referees for their suggestions which helped improved this paper.

## References

Catalisano, M.V., Geramita, A. V., Gimigliano, A. (2002). Ranks of tensors, secant varieties of Segre varieties and fat points, *Linear Algebra and its Applications* **355**, pp. 263–285.

Cox, D., Little, J. and O'Shea, D. (1997). *Ideals, Varieties and Algorithms*, Springer Undergraduate Texts in Mathematics, Second Edition.

Garcia, L. D., Stillman, M. and Sturmfels, B. (2004). Algebraic Geometry of Bayesian Networks, *Journal of Symbolic Computation*, accepted.

Geiger, D., Heckerman, D., King, H. and Meek, C. (2001). Stratified exponential families: graphical models and model selection, *Annals of Statistics* **29**, pp. 505–529.

Geiger, D. and Meek, C. (1998). Graphical models and exponential families, *Proceedings of the Fourteenth Annual Conference on Uncertainty in Artificial Intelligence*, pp. 156–165, San Francisco, CA. Morgan Kaufmann Publishers.

Goodman, L. (1974). Explanatory latent structure analysis using both identifiable and unidentifiable models, *Biometrika* **61**, pp. 215–231.

Harris, J. (1992). *Algebraic Geometry*, Springer Graduate Texts in Mathematics.

Haughton, D. (1988). On the choice of a model to fit data from an exponential family, *Annals of Statistics* **16**, pp. 342–555.

Kočka, T. and Zhang, N. L. (2002). Dimension correction for hierarchical latent class models, *Proceedings of the Eighteenth Conference on Uncertainty in Artificial Intelligence (UAI-02)*.

Landsberg, J. M. and Manivel, L. (2003) On the ideals of secant varieties of Segre varieties, submitted.

Lauritzen, S. L. (1996). *Graphical Models*, Oxford University Press.

Pistone, G., Riccomagno, E. and Wynn, H. (2001). *Algebraic Statistics: Computational Commutative Algebra in Statistics*, Chapman and Hall, Boca Raton.

Rusakov, D. and Geiger, D. (2002). Asymptotic model selection for naive Bayesian networks, *Proceedings of the Eighteenth Annual Conference on Uncertainty in Artificial Intelligence (UAI-02)*.

Rusakov, D. and Geiger, D. (2003). Automated analytic asymptotic evaluation of the marginal likelihood for latent models, *Proceedings of the Nineteenth Annual Conference on Uncertainty in Artificial Intelligence (UAI-03)*.

Settimi, R. and Smith, J. Q. (2000). Geometry, moments and conditional independence trees with hidden variables, *Annals of Statistics* **28** pp. 1179–1205.

Settimi, R. and Smith, J. Q. (1998). On the geometry of Bayesian graphical models with hidden variables, *Proceedings of the Fourteenth Conference on Uncertainty in Artificial Intelligence*, pp. 472–479, San Francisco, CA. Morgan Kaufmann Publishers.

Schwarz, G. (1978). Estimating the dimension of a model, *Annals of Statistics* **6(2)** pp. 461–464.

Spirtes, P., Glymour, C. and Scheines, R. (1993). *Causation, Prediction, and Search*. Springer–Verlag.

Terracini, A. (1911). Sulle $V_K$ per cui la varietà degli $S_h(h+1)$-seganti ha dimensione minore dell'ordinario, *Rend. Circ. Mat. Palermo* **31 pp. 392–396.**